\documentclass[letterpaper, 10 pt, conference]{ieeeconf}  

\IEEEoverridecommandlockouts                              

\usepackage{float}
\usepackage{cite}
\usepackage{amsmath,amssymb,amsfonts}
\usepackage{gensymb}
\usepackage{graphicx}
\usepackage{textcomp}
\usepackage{xcolor}
\usepackage{verbatim}
\usepackage{afterpage}
\usepackage{cite}
\usepackage{overpic}
\usepackage{psfrag}
\usepackage{latexsym}
\usepackage{bm}
\usepackage{cases}
\usepackage{array}
\usepackage{fancyhdr}
\usepackage{setspace}
\usepackage{subfigure}
\usepackage{url}
\usepackage{multirow}
\usepackage{epstopdf}
\usepackage{epsfig}
\usepackage{fancybox}
\usepackage{textcomp}
\usepackage{physics}
\usepackage{romannum}
\usepackage{caption}
\usepackage{float}
\usepackage[ruled,linesnumbered]{algorithm2e}

\DeclareMathOperator*{\argminA}{arg\,min} 
\newcommand{\x}[0]{\mathbf{x}}

\newcommand{\R}[0]{\mathbb{R}}

\def\BibTeX{{\rm B\kern-.05em{\sc i\kern-.025em b}\kern-.08em
    T\kern-.1667em\lower.7ex\hbox{E}\kern-.125emX}}
\begin{document}

\title{Moving Object Localization based on the Fusion of Ultra-WideBand and LiDAR with a Mobile Robot
\thanks{M. Shalihan, Z. Cao, K. Pongsirijinda, B. P. L. Lau, R. Liu, and U-X. Tan are with the Engineering Product Development Pillar, Singapore University of Technology and Design, 8 Somapah Rd, Singapore, 487372. {\{\tt\small muhammad\_shalihan,zhiqiang\_cao, khattiya\_pongsirijinda\}@mymail.sutd.edu.sg} {\{\tt\small billy\_lau@sutd.edu.sg,ran
\_liu,uxuan\_tan\}
\\@sutd.edu.sg} 
\\C. Yuen is with the School of Electrical \& Electronic Engineering, Nanyang Technological University, 50 Nanyang Ave, 639798. {\{\tt\small chau.yuen\}@ntu.edu.sg} 
\\L. Guo is with the Southwest University of Science and Technology, Mianyang, Sichuan, China 621010.
}
}

\author{Muhammad Shalihan, Zhiqiang Cao, Khattiya Pongsirijinda, Lin Guo, Billy Pik Lik Lau, Ran Liu, Chau Yuen, \\and U-Xuan Tan
}

\maketitle

\begin{abstract}
Localization of objects is vital for robot-object interaction. Light Detection and Ranging (LiDAR) application in robotics is an emerging and widely used object localization technique due to its accurate distance measurement, long-range, wide field of view, and robustness in different conditions. However, LiDAR is unable to identify the objects when they are obstructed by obstacles, resulting in inaccuracy and noise in localization. To address this issue, we present an approach incorporating LiDAR and Ultra-Wideband (UWB) ranging for object localization. The UWB is popular in sensor fusion localization algorithms due to its low weight and low power consumption. In addition, the UWB is able to return ranging measurements even when the object is not within line-of-sight. Our approach provides an efficient solution to combine an anonymous optical sensor (LiDAR) with an identity-based radio sensor (UWB) to improve the localization accuracy of the object. Our approach consists of three modules. The first module is an object-identification algorithm that compares successive scans from the LiDAR to detect a moving object in the environment and returns the position with the closest range to UWB ranging. The second module estimates the moving object's moving direction using the previous and current estimated position from our object-identification module. It removes the suspicious estimations through an outlier rejection criterion. Lastly, we fuse the LiDAR, UWB ranging, and odometry measurements in pose graph optimization (PGO) to recover the entire trajectory of the robot and object. For a static robot and a moving object scenario, we show in experiments that the proposed approach improves the average relative translational and rotational accuracy by 44\% and 31.6\%, respectively, compared to the conventional UWB ranging localization. Additionally, we extend the approach to a moving robot and a moving object scenario and show that our approach improves the average relative translation and rotational accuracy by 13.5\% and 36\%, respectively.    

\end{abstract}

\section{Introduction}
Object localization is essential for many applications \cite{guo2019survey}. The literature shows several approaches for object localization in environments with a prior map or known infrastructure \cite{dong2018vinav}. For example, the Global Positioning System (GPS) can provide meter-level accuracy but is unsuitable for indoor applications due to blocked signals by surrounding buildings \cite{li2018gps}. However, some applications may not have prior information on the environment. Therefore, in these environments, localization between a robot and an object is crucial to accomplish a number of tasks, such as object-following scenarios where a robot needs to follow a moving object. Various sensors can be utilized for object localization, such as odometry, Inertial Measurement Unit (IMU), Light Detection and Ranging (LiDAR), visual cameras, and Ultra-wideband (UWB) sensors. Odometry through wheel encoders or the IMU is commonly used for localization as it provides an estimated position with reference to the starting position. The odometry shows good accuracy when used for short periods. However, odometry is commonly known to drift over-time \cite{wheelslippage}. Similarly, the IMU can measure position changes but deteriorates over time due to the accumulation of error \cite{liu2020cooperative}. Therefore, localization has been extensively researched with the use of different sensors to improve the localization accuracy of objects \cite{liu2022distributed}.

Approaches for localization using LiDAR and visual-based sensors such as the monocular camera provide accurate localization results \cite{tian2022kimera}. With this, LiDAR and visual-based sensors are beneficial for object detection when in line-of-sight. However, it is difficult to track the object of interest from the rest of the obstacles and moving objects in the environment. This difficulty arises because the LiDAR and visual-based sensors can only distinguish between different objects by applying computer vision approaches such as \cite{cv} or the help of additional sensors such as the UWB sensor. The UWB is popularly used to improve localization results \cite{yassin2016recent}\cite{liu2017cooperative}. The UWB provides up to centimetre-level ranging accuracy under line-of-sight conditions. However, the UWB does not provide the bearing information and is susceptible to multi-path measurements under non-line-of-sight conditions \cite{shalihan2022nlos}. Non-line-of-sight UWB measurement mitigation approaches, such as in \cite{shalihan2022nlos}, require data collection and training of the neural network model before application, which could take up much time. Therefore, we focus on the pose estimation of a robot and a moving object through odometry, UWB ranging, and LiDAR measurements. 

\begin{figure}
\centering
\includegraphics[width=0.435\textwidth]{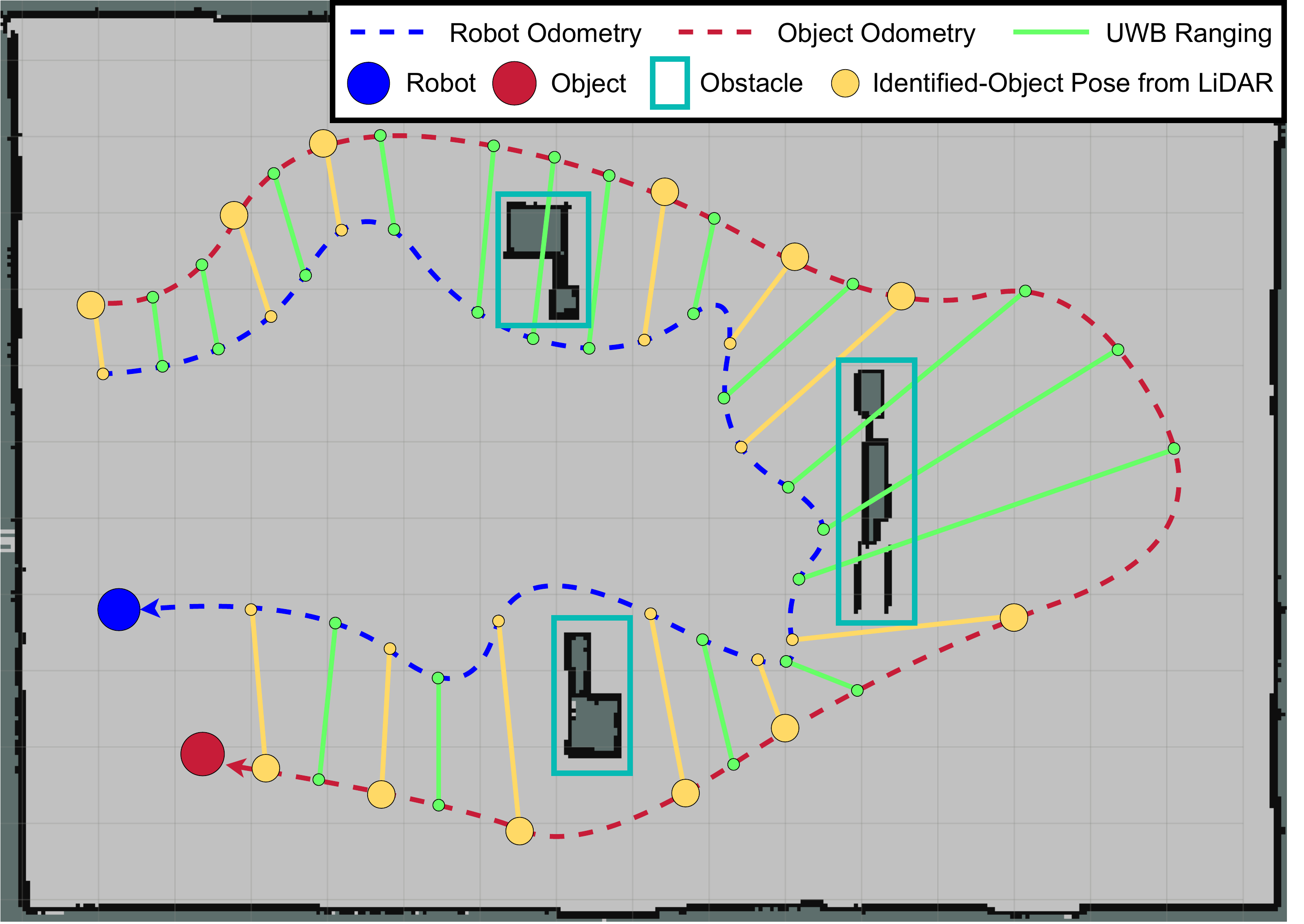}
\caption{Overview of the proposed localization approach with a robot (blue colour) and a moving object (red colour). We perform pose estimation with UWB ranging (green lines) and odometry (red and blue dotted lines) measurements, as well as LiDAR measurements produced by our object-identification module (see yellow points). The estimated poses from the object-identification module are passed for outlier rejection and optimized through a pose graph optimization module. Mapped obstacles are for visualization purposes only.
}
\vspace{-0.5cm}
\label{fig:overview}
\end{figure}

The main idea of our approach is to improve the localization results of current UWB-ranging localization approaches by incorporating LiDAR measurements as a constraint through a pose graph optimization (PGO) framework. In our experiments, a moving object has both odometry and UWB data. The robot provides odometry, carries a UWB for ranging, and also carries a 2D-LiDAR to identify the moving object. A simple illustration of the proposed method can be found in Figure \ref{fig:overview}, where the map is only used for visualization purpose only, as our approach targets real-life scenarios where a prior map of the environment is not available   

We propose fusing UWB ranging, odometry, and LiDAR measurements in line-of-sight conditions to produce accurate pose estimations. Even though the LiDAR measurements are only available during line-of-sight scenarios, the overall trajectory of the object will be improved given accurate object-identification pose estimates. To eliminate false positives, we introduce a heuristic outlier rejection mechanism that compares the estimation of the object's moving direction from our module and the current estimated object's orientation from PGO. Finally, inlier LiDAR measurements are fused with UWB ranging and odometry through PGO to estimate the object's trajectory. 
The contributions of this paper are summarized as follows:
\begin{itemize}
\item We propose an approach for pose estimation between a robot and an object using UWB ranging, odometry, and LiDAR measurements. In particular, an object-identification module is used to identify the object through LiDAR scans and improve localization accuracy using the detected object position compared to the conventional UWB ranging localization. 
\item We present an approach to determine the moving direction of an object based on successive LiDAR scans to improve localization accuracy and introduce a mechanism to reject incorrect measurements by comparing results from the object-identification module with estimated pose from optimization.
\item We perform extensive experiments to evaluate the performance of the proposed approach in a complex environment of size 16m$\times$12m. We have achieved an improvement of 44\% and 31.6\% in translational and rotational accuracy, respectively, for a static robot and a moving object scenario. For the moving robot and moving object scenario, we achieve an improvement of 13.5\% and 36\%, respectively, in translation and rotational accuracy.
\end{itemize}

We organize the remaining of this paper as follows: Section II introduces the related work. Section III describes the proposed localization technique to fuse UWB ranging, odometry, and LiDAR measurements. Section IV shows the experimental setups and results. Finally, Section V concludes this paper and discusses future work.

\section{Related work}
The robotics community shows a growing interest in object localization, especially for real-world applications such as when a robot needs to follow an object. As a result, there is growing research on object localization through different approaches. For example, Long \emph{et. al.} \cite{long2017accurate} proposed a method for accurate object localization by introducing three processes: region proposal, classification, and accurate object localization. Features extracted from a region of interest from an image through a convolutional neural network go through an unsupervised bounding box regression algorithm that localizes and optimizes the position of the detected object. Similarly, Tychsen-Smith \emph{et. al.} \cite{tychsen2018improving} proposed a similar approach but designed the Fitness Non-Max Supression (NMS) and derived a novel bounding box regression based on a set of Intersection-over-Union (IoU) upper bounds to obtain greater localization accuracy. However, the methods mentioned above require training a model. Although these methods can be implemented on smaller training data, the performance is unsatisfactory due to limitations in feature representation and model complexity \cite{zhang2021weakly}. 

Instead of processing images from a visual camera with deep-learning methods, point cloud data from the LiDAR can provide high positioning accuracy for object localization. For example, Huang \emph{et. al.} \cite{2} proposed a frame-to-frame scan matching algorithm based on an attention mechanism. In this method, the selected landmark is not switched to another before it becomes invisible. Therefore, the approach will not accumulate errors while the landmark is not changed, giving high matching accuracy. Successive scans can then be compared, similar to the work of Mih\'alik \emph{et. al.} \cite{3} to identify moving objects based on Euclidean distance moved by the points in the point cloud. However, with multiple moving objects in the environment, an additional sensor, such as the UWB, may be required to accurately narrow down the results to identify the object of interest. The UWB is popular among the robotics community to help compensate for odometry errors caused by drift over long periods due to its low cost, and low power level consumption. For example, Liu \emph{et. al.} \cite{liu2022distributed} and Cao \emph{et. al.} \cite{cao2021relative} improve pose estimation results by minimizing UWB rangings taken at different positions and fusing the UWB ranging measurements with odometry through PGO. In addition to the UWB helping in object-identification, and improving pose estimations in PGO, the LiDAR measurements from the object-identification can also be fused with the UWB ranging measurements to improve results further.

Research on the fusion of UWB ranging and LiDAR measurements have been performed with promising results. Song \emph{et. al.} \cite{5} and Zhou \emph{et. al.} \cite{6} proposed a UWB/LiDAR fusion for cooperative range-only SLAM. Building on the work of Ding \emph{et. al.} \cite{7} which employs LiDAR Inertial Odometry (LIO) for robust LiDAR localization, Nguyen \emph{et. al.} \cite{8} proposed the LiDAR-Inertia-Ranging Odometry (LIRO) localization by introducing UWB ranging measurements for fusion with LiDAR and inertial measurements. The LIRO approach improves localization accuracy compared to LIO by having only two or three anchors deployed in the environment. With the LiDAR giving a more comprehensive picture of the surrounding environment, fusing LiDAR measurements with the UWB ranging measurements helps to remove errors accumulated in the LiDAR-based SLAM algorithms. However, the experiment mentioned above includes UWB beacons placed around the environment to provide ranging measurements between robots and nearby obstacles, which is not ideal, especially in emergencies. We propose a method to fuse odometry, UWB ranging, and LiDAR measurements in a PGO without the need for infrastructure.

\section{Localization between a robot and an object}
\begin{figure}[h]
\centering
\includegraphics[width=0.4\textwidth]{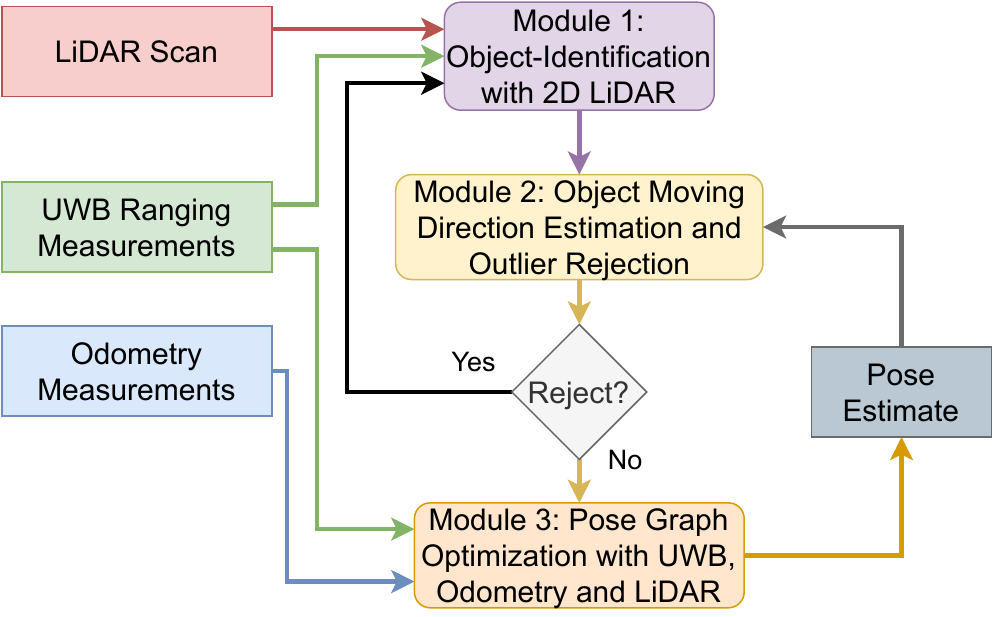}
\caption{Flowchart of the proposed object localization approach by fusing odometry, UWB ranging, and LiDAR measurements. 
}
\label{fig:flowchart}
\vspace{-0.25cm}
\end{figure}

In this section, we formulate the problem for localization between a robot and an object using UWB ranging, odometry and LiDAR measurements without prior knowledge about the infrastructure. An overall view of our proposed approach is shown in Figure \ref{fig:flowchart}. It consists of three modules, which are (1) Object-Identification with 2D LiDAR, (2) Object Moving Direction Estimation and Outlier Rejection, and (3) Pose Graph Optimization with UWB, Odometry, and LiDAR. First, we identify the object's pose based on the LiDAR measurements from the robot in the object-identification module. Next, we compute the moving direction of the object and perform outlier rejection by comparing the object's pose output from our object-identification algorithm with the current estimated pose from PGO through a heuristic strategy. Lastly, we estimate the robot's and object's pose by incorporating UWB ranging, odometry, and the object's pose identified by our object-identification module through PGO.

\subsection{Object-Identification with 2D LiDAR}
\begin{algorithm}

\label{particle_filter_algo}

\SetKw{KwWith}{with}

\SetKw{KwEach}{each}

  \SetKwData{Left}{left}\SetKwData{This}{this}\SetKwData{Up}{up}
  \SetKwFunction{Union}{Union}\SetKwFunction{FindCompress}{FindCompress}
  \SetKwInOut{Input}{input}\SetKwInOut{Output}{output}
  \SetKwInput{KwData}{Input}
  \SetKwInput{KwResult}{Output}

\KwData{\small LiDAR point cloud with reference to the robot $\mathbf{P}_{R}^t$ at time $t$ and UWB ranging between the robot and object: $r^t$ at time $t$.} 

\KwResult{\small Object pose estimation from LiDAR at time $t$ $\overline{\x}_L^t$.}

\caption{\small Estimate object pose at time $t$ $\overline{\x}_L^t$ from LiDAR with reference to the  robot.}

\tcp{\small Adaptive Clustering of $\mathbf{P}_{R}^t$ according to \cite{adaptivecluster} returns array of clusters $C^{t}$.}
  $\rhd$ \small Compare $C^{t-1}$ with $C^{t}$ based on Euclidean distance and return array of moving objects $C_{dynamic}^{t}$ at time $t$ according to \cite{3}.\
\\ \tcp{\small Narrow down to the position of interest from the cluster array of moving objects.}  
    \For{\small$C_{k}^t$ $\in$ $C_{dynamic}^{t}$}
    {
   $\rhd$ \small Compute Euclidean distance of $C_{k}^t$ $\in$ $C_{dynamic}^{t}$ with respect to the  robot. \\
   $\rhd$ \small Cluster position with a distance to the robot closest to the current UWB ranging $r^t$ is estimated as the object 2D position $\x_L^t$.
    }
 $\rhd$ \small Compute the estimated moving direction of the object $\theta_L^t$ at time $t$ with 2D positions $\x_{L}^t$ and $\x_L^{t-1}$.
 
 $\rhd$ \small Return the estimated object pose identified by LiDAR  $\overline{\x}_L^t$ at time $t$ given by the 2D the position $\x_L^t$ and moving direction $\theta_L^t$.  
\end{algorithm}
An overview of the object-identification module is shown in Algorithm 1. The pose of the object obtained through our object-identification and object moving-direction estimation module at time $t$ is denoted as $\overline{\x}_L^t=[\x_L^t,\theta_L^t]$, where $\x_L^t$ is the 2D position and $\theta_L^t$ is the heading. The subscript $L$ indicates that the estimation is from our proposed object-identification and object-moving-direction estimation module. The laser scans from the 2D LiDAR with reference to the robot are converted into a 2D point cloud $\mathbf{P}_{R}^t$ = $\{p_{i} | p_{i} = (x_{i},y_{i}) \in \R^2, i = 1, ..., I\}$, where \emph{I} is the total number of points in a single scan. Next, adaptive clustering is performed according to \cite{adaptivecluster}. Adaptive clustering was chosen instead of the conventional clustering algorithm \cite{9072123}. Instead of a fixed-distance threshold, which can be inaccurate, the adaptive clustering algorithm will return the pose of the segmented objects centre based on a Euclidean distance threshold:

\begin{equation}
    d_{i}^* = 2 \cdot cr_{i} \cdot \tan \frac{\Theta}{2},
    \label{distancethreshold}
\end{equation}

\noindent where $\Theta$ refers to the angular resolution of the LiDAR. A set of values $d_{i}^*$ are considered at fixed intervals to compute the maximum cluster range $cr_{i}$ using the inverse of Equation \ref{distancethreshold}. This returns cluster positions that include the object's position with reference to the robot in a point cloud array $C^{t}$ = $\{c_{k}^t | c_{k}^t = (x_{k}^t,y_{k}^t) \in \R^2, k = 1, ..., K\}$ at time $t$, where \emph{K} is the total number of clusters in the point cloud $\mathbf{P}_{R}^t$.

The position of the object $\x_L^t$ can be estimated by comparing two successive point clouds from the LiDAR. This will result in an array of positions from all moving objects, including all the ones not intended. Therefore, we narrow down to the object of interest from the array of moving objects ($C_{dynamic}^{t}$) using UWB ranging. First, we compute the distances between the robot and the different positions $C_{dynamic}^{t}$. Next, we compare the difference between UWB ranging and the computed distances. The position from $C_{dynamic}^{t}$ with the closest distance to the current UWB-range measurement and within 0.3m of the current UWB-range is identified as the object's pose from the LiDAR. Due to the criterion employed here, false positives are minimized. This estimated position is used as the input for our object-moving-direction estimation and outlier rejection module, which will be explained in the next subsection.

\subsection{Object Moving Direction Estimation and Outlier Rejection}
In this module, we estimate the object's moving direction using two successive object 2D positions, $\x_L^{t-1}$ and $\x_L^{t}$, from our object-identification module. The estimated object moving direction is given by:


\begin{equation}
 \theta_L^t = \operatorname{atan2}(y_L^{t}-y_L^{t-1},x_L^{t}-x_L^{t-1}),
\label{theta_estimate}
\end{equation}

\noindent where $\x_L^{t-1} = [x_L^{t-1},y_L^{t-1}]$ denotes the previous estimated object 2D position and $\x_L^{t} = [x_L^{t},y_L^{t}]$ denotes the current estimated object 2D position from the LiDAR in our proposed object-identification algorithm.

As there may be multiple misidentified objects from the object-identification module due to multiple dynamic obstacles in the environment, we introduce a heuristic outlier rejection mechanism to remove suspicious moving direction measurements estimated by our object-identification module in the previous subsection. Given the estimated moving direction from our object-identification module $\theta_L^t$, we compare it with the current estimated object orientation from PGO $\theta_O^{t}$ as follows: 

\begin{equation}
    \Omega_{L,\theta}^{t}= 
\begin{cases}
    \omega,& \text{if } |\theta_L^t-\theta_O^{t}| \leq \vartheta\\
    0,              & \text{otherwise}
\label{heuristic}
\end{cases},
\end{equation}

\noindent where $\vartheta$ is the error threshold set in radians, and $\omega$ is the value set for the LiDAR moving direction information matrix $\Omega_{L,\theta}^{t}$. If the condition is satisfied, the moving direction information matrix value will be set to a high value $\omega$ to indicate that the measurement is trusted. If the condition is not satisfied, the moving direction information matrix will be set to 0 to indicate that the measurement cannot be trusted. It is unlikely for the orientation of the moving object to change drastically from its previous orientation. Therefore, a threshold value to reject false positives using the currently estimated object orientation from PGO works well and does not break the estimation due to a fast-moving orientation change. PGO is then performed after rejecting outlier moving direction measurements to improve localization accuracy. The parameters $\vartheta$ and $\omega$ will be further studied in the next section.

\subsection{Pose Graph Optimization using UWB, Odometry, and LiDAR}
\label{pose_est_ranging}
The objective is to estimate the trajectory of the robot $\overline{\x}_{R}^{1:T}$=$\{\overline{\x}_{R}^{1}$,..., $\overline{\x}_{R}^{T}\}$ and the object $\overline{\x}_{O}^{1:T}$=$\{\overline{\x}_{O}^{1}$,..., $\overline{\x}_{O}^{T}\}$ from time 1 up to time $T$, where $\overline{\x}_{O}^{t}$=$[x_{O}^{t}$, $y_{O}^{t}$, $\theta_{O}^{t}]$  and $\overline{\x}_{R}^{t}$=$[x_{R}^{t}$, $y_{R}^{t}$, $\theta_{R}^{t}]$ represents the pose of the object and robot obtained through PGO at time $t$ respectively. We use $r^{t}$ to denote the UWB ranging between the robot and the object at time $t$. The LiDAR measurements, which include the object's 2D position and moving direction obtained through our object-identification and object-moving-direction estimation module at time $t$, are denoted as $\x_L^t$ and $\theta_L^t$ respectively. We estimate the robot and object pose given the odometry, UWB ranging, and LiDAR measurements through a centralized PGO solution. The PGO technique uses the poses of the robot and objects as nodes in a graph to be estimated. Each node represents a specific pose at a given time step. The graph's edges connect these nodes based on the measurements provided, forming constraints between them. To find the optimal configuration of poses, PGO minimizes the error of these constraints using methods such as maximum likelihood estimation or nonlinear optimization techniques. This process leads to a refined and globally consistent estimation of the poses in the system's trajectory. In our approach, the errors to be minimized through maximum likelihood estimation is as followed:

\begin{equation}
\small
 \begin{split}
\argminA_{\overline{\x}_{R}^{1:T},\overline{\x}_{O}^{1:T}}&\sum_{t=2}^{T} \underbrace{\mathbf{e}(\overline{\x}_{R}^{t-1}, \overline{\x}_{R}^{t}, \Delta\overline{\x}_{R}^{t})^T\Omega_{R}^{t}\mathbf{e}(\overline{\x}_{R}^{t-1}, \overline{\x}_{R}^{t}, \Delta\overline{\x}_{R}^{t})}_{\text{Robot odometry constraint}} {+} \\
&\sum_{t=2}^{T} \underbrace{\mathbf{e}(\overline{\x}_{O}^{t-1}, \overline{\x}_{O}^{t}, \Delta\overline{\x}_{O}^{t})^T\Omega_{O}^{t}\mathbf{e}(\overline{\x}_{O}^{t-1}, \overline{\x}_{O}^{t}, \Delta\overline{\x}_{O}^{t})}_{\text{Object odometry constraint}} {+} \\
&\sum_{t=1}^{T} \underbrace{\mathbf{e}(\overline{\x}_{R}^{t}, \overline{\x}_{O}^{t}, r^t)^T\Omega_{R, O}^{t}\mathbf{e}(\overline{\x}_{R}^{t}, \overline{\x}_{O}^{t}, r^t)}_{\text{UWB ranging constraint}} {+} \\
&\sum_{t=1}^{T} \underbrace{\mathbf{e}(\overline{\x}_{R}^{t}, \overline{\x}_{O}^{t}, \x_{L}^t)^T\Omega_{L,\x}^{t}\mathbf{e}(\overline{\x}_{R}^{t}, \overline{\x}_{O}^{t}, \x_{L}^t)}_{\text{Object position constraint (Module 1)}} {+} \\
&\sum_{t=1}^{T} \underbrace{\mathbf{e}(\overline{\x}_{R}^{t}, \overline{\x}_{O}^{t}, \theta_{L}^t)^T\Omega_{L,\theta}^{t}\mathbf{e}(\overline{\x}_{R}^{t}, \overline{\x}_{O}^{t}, \theta_{L}^t)}_{\text{Object moving direction constraint (Module 2)}}, 
\label{eq:optimization}
 \end{split}
\end{equation}

\noindent where $\mathbf{e}(\cdot)$ denotes the residual function that computes the residual error of odometry, UWB ranging and a LiDAR pose measurement from our object-identification module given a pose configuration for the robot $\overline{\x}_{R}^t$ and the object $\overline{\x}_{O}^t$. Constraints are additionally
parameterized with a certain degree of uncertainty,
which is denoted as the information matrix (i.e., $\Omega_{R}^{t}$,$\Omega_{O}^{t}$,$\Omega_{R, O}^{t}$,$\Omega_{L,\x}^{t}$,$\Omega_{L,\theta}^{t}$) in Equation \ref{eq:optimization}.

Due to the non-convexity of Equation \ref{eq:optimization}, the optimization converges to a local minimum without a reasonable initial guess, and there is no guarantee of finding the best solution. Therefore, we use the known initial robot and object position with reference to the robot as an initial guess for optimization. Based on the initial guess, we then perform nonlinear optimization via g2o \cite{22}. Next, we study all three modules using real-world experiments to show their effectiveness. 

\section{Experimental results}
In this section, we present the experimental results of our approach and compare them with two other methods: pure odometry and conventional UWB ranging localization. Furthermore, we include results for the fusion of odometry and LiDAR to provide a comprehensive comparison. We begin by describing the experimental setups in Section IV-A. Subsequently, we present the results obtained using a setup comprising a static robot equipped with a 2D LiDAR and a moving object (robot without LiDAR), and evaluate the pose estimation outcomes in Section IV-B. Building upon that, we extend the proposed approach to a setup involving a moving robot (with LiDAR) and a moving object (robot without LiDAR), and assess the performance of our method in Section IV-C.

The first experiment focuses on verifying the effectiveness of the proposed approach using a static robot and a moving object. Additionally, in this experiment, we identify the optimal values for $\vartheta$ and $\omega$. The second experiment aims to validate the robustness of our approach on a moving robot and a moving object, utilizing the best values for $\vartheta$ and $\omega$ based on the previous experiment. All experiments were conducted on a system equipped with an Intel Core i7-6600U CPU for reliable and consistent performance.

\subsection{Experiment Setups}
\begin{figure}[h]
\centering
\includegraphics[width=0.485\textwidth]{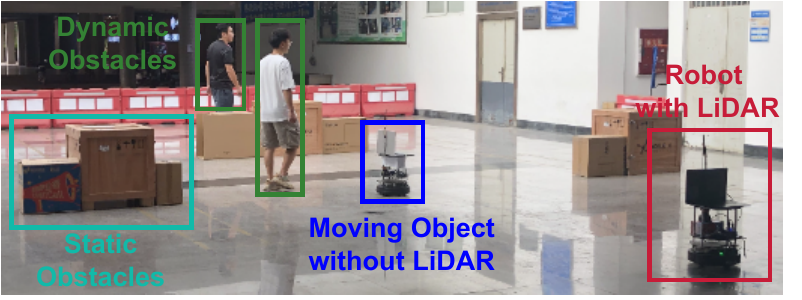}
\caption{Snapshot of experiment setup with a  robot (turtlebot equipped with LiDAR) and moving object (turtlebot without LiDAR) in an indoor environment with static and dynamic obstacles.}
\label{fig:robot}
\vspace{-0.3cm}
\end{figure}

\begin{figure*}
  \centering
  \subfigure[Pose estimation of the moving object with respect to the static robot using only odometry measurements.]{
\label{figure:track_raw}
        \includegraphics[width=0.3\textwidth]{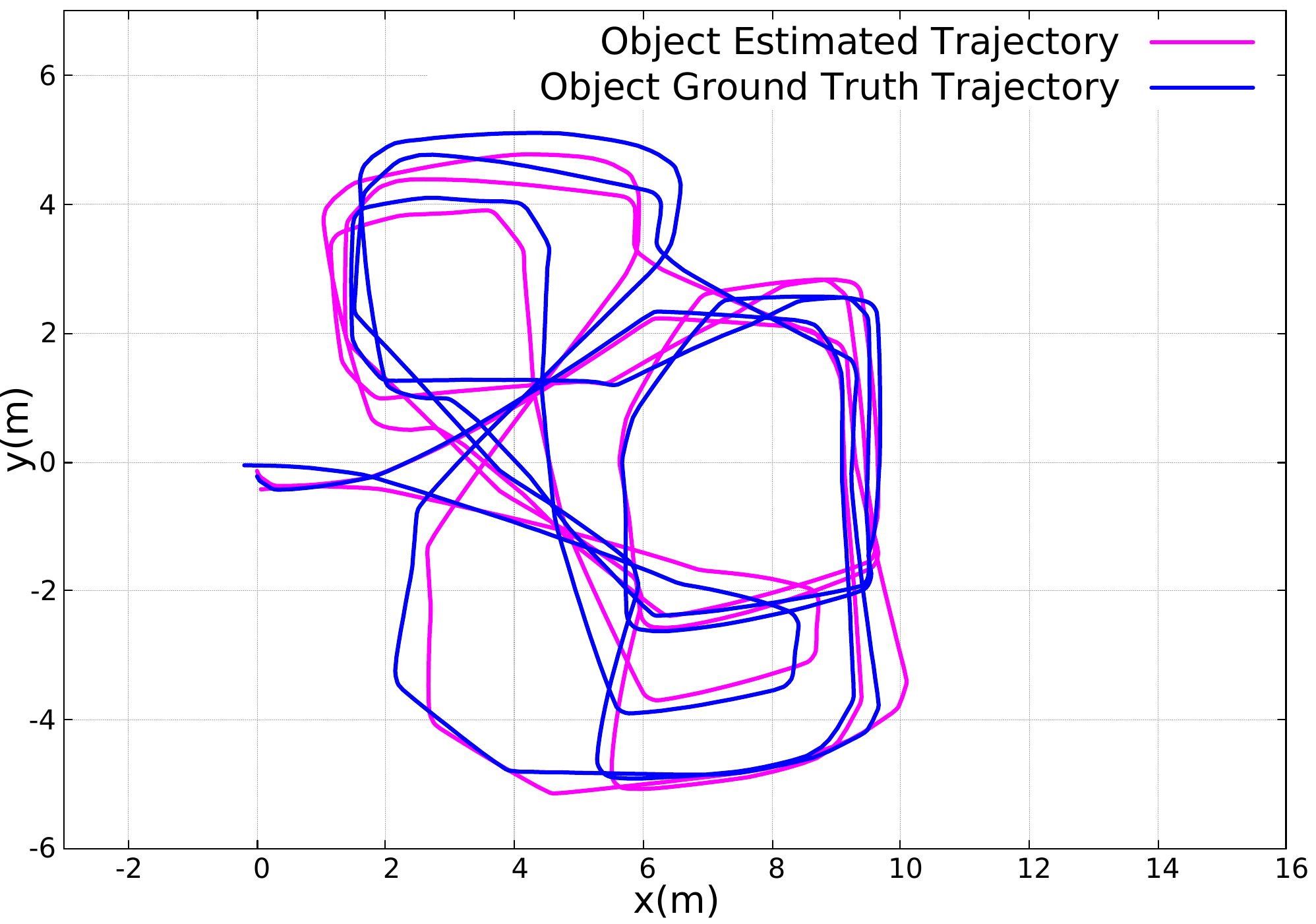}
        }\hspace{0.2in}\subfigure[Pose estimation of the moving object with respect to the static robot using UWB ranging and odometry measurements.]{
\label{figure:track_pcm}
        \includegraphics[width=0.3\textwidth]{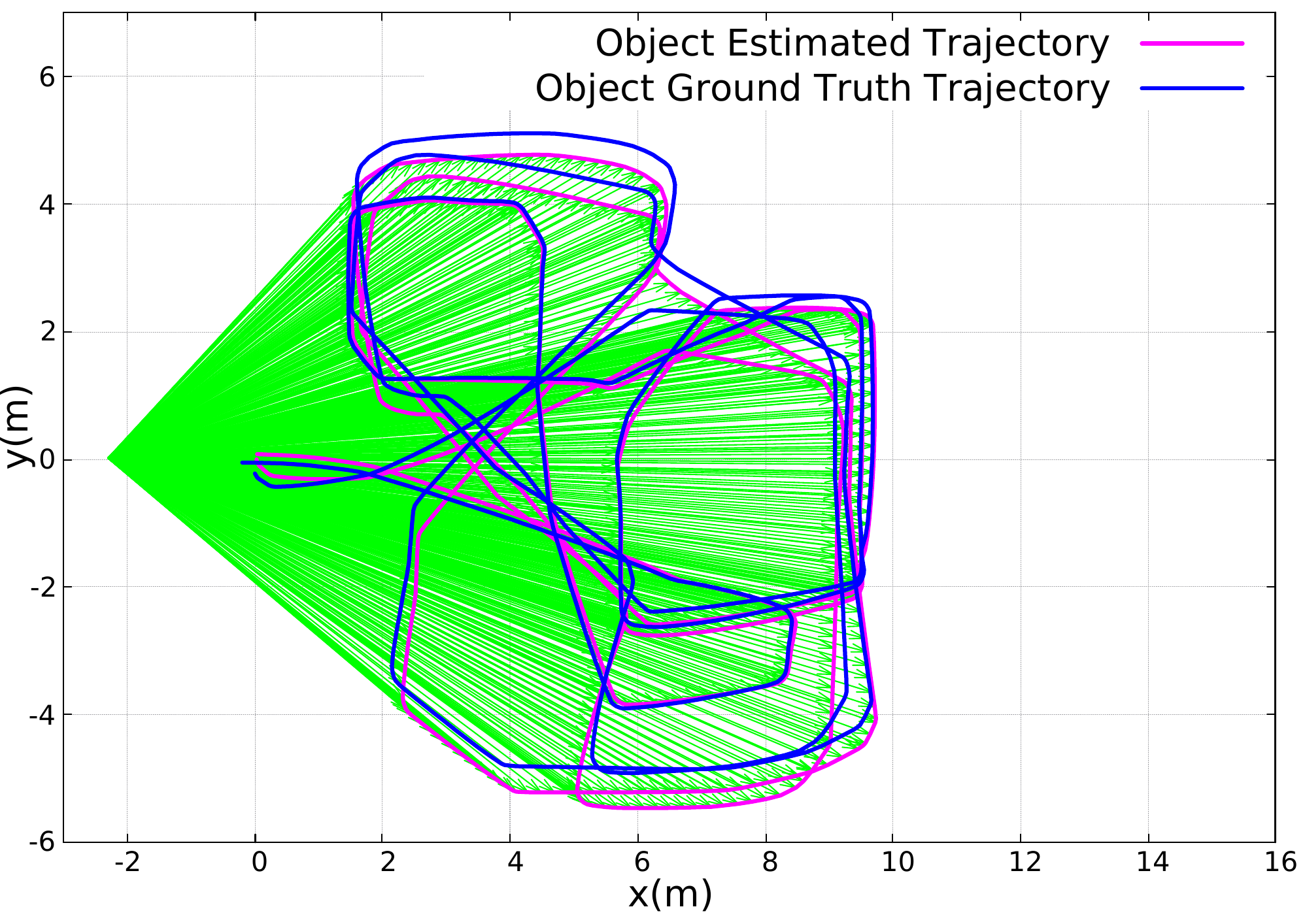}
        }\hspace{0.2in}\subfigure[Pose estimation of the moving object with respect to the static robot using UWB ranging, odometry and LiDAR measurements (with both moving direction and rejection).]{
\label{figure:track_pgo}
        \includegraphics[width=0.3\textwidth]{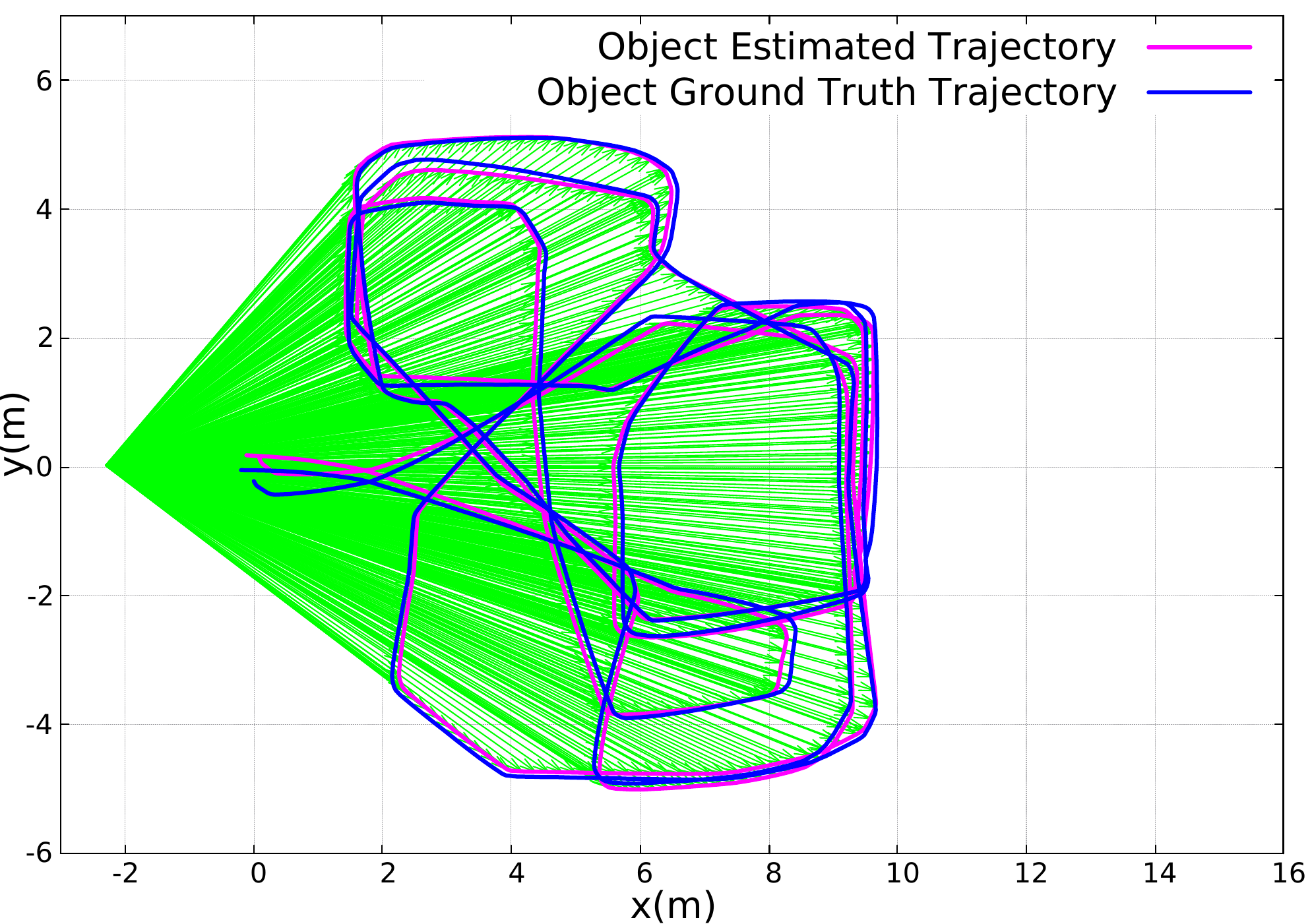}
        }
   \caption[]
{Trajectories estimated by different approaches for the static robot and a moving object scenario. The green lines denote the UWB ranging constraints, and the point of intersection for the green lines is the static robot.}
\label{figure:trajectory}
\vspace{-0.3cm}
\end{figure*}

In this subsection, we present the experimental setup to demonstrate the proposed approach with a  robot (a turtlebot with LiDAR) and an object (a turtlebot without LiDAR) in an indoor environment of 16m$\times$12m which consists of static and dynamic obstacles. Fig. \ref{fig:robot} shows a snapshot of the experimental setup. The robot and object each carry one UWB node (NoopLoop LinkTrack) with a range of up to 100m and a sampling frequency of 50Hz. The robot and object output odometry measurements with a frequency of 10Hz. In addition, the robot also carries a 2D LiDAR (Hokuyo LiDAR), which publishes laser scans at 20Hz. All modules are run using the Robot-Operating-System (ROS) \cite{ros}.

For the experiments, ground truth was obtained using the Hokuyo LiDAR to perform Adaptive Monte Carlo Localization (AMCL) \cite{AMCL} given a map created through GMapping \cite{Gmapping}, which provides accurate pose estimations for both the robot and the object. The accuracy of our proposed approach was evaluated by comparing the computed relative translational and rotational errors of the robot and the object against the ground truth data. 

\subsection{Experiments with a Static Robot and a Moving Object}
This section presents the experimental results to demonstrate the proposed approach with a static robot and a moving object. The robot is static at its initial position, while the moving object moves along different paths.
\subsubsection{Evaluation of Pose Estimation using UWB, Odometry and LiDAR}
We commence our evaluation by examining pure odometry, a method known to exhibit drift over time. Next, we assess UWB ranging localization \cite{cao2021relative}, which combines UWB ranging and odometry measurements to minimize the residual error of the UWB range. The third approach under evaluation is the fusion of odometry and LiDAR measurements, without the incorporation of UWB ranging as constraints. Lastly, we evaluate our proposed approach, which integrates LiDAR, UWB ranging, and odometry measurements for accurate pose estimation. Furthermore, we investigate the impact of the LiDAR moving direction information matrix value on the precision of the pose estimation results.
\begin{table}[h]
\centering
\caption{Evaluation of different approaches based the average relative translational error (metres) and relative rotational error (radians) between the static robot and the moving object.}
\label{results}
\resizebox{\linewidth}{!}{%
\begin{tabular}{|c|c|c|}
\hline
Approach                                                                                        & Rel. trans error (m) & Rel. rot error (rad) \\ \hline
Pure Odom                                                                                       & 0.28 ± 0.12          & 0.066 ± 0.024        \\ \hline
Odom + UWB                                                                                      & 0.25 ± 0.092         & 0.038 ± 0.011        \\ \hline
\begin{tabular}[c]{@{}c@{}}Odom + LiDAR\\ (with moving direction\\ and rejection)\end{tabular}  & 0.26 ± 0.056         & 0.039 ± 0.009        \\ \hline
\begin{tabular}[c]{@{}c@{}}Odom + UWB + LiDAR\\ (w/o moving direction)\end{tabular}                  & 0.29 ± 0.15          & 0.061 ± 0.031        \\ \hline
\begin{tabular}[c]{@{}c@{}}Odom + UWB + LiDAR\\ (w/o rejection)\end{tabular}                    & 0.23 ± 0.070         & 0.057 ± 0.024        \\ \hline
\begin{tabular}[c]{@{}c@{}}Odom + UWB + LiDAR\\ (with moving direction\\ and rejection)\end{tabular} & 0.14 ± 0.032         & 0.026 ± 0.006        \\ \hline
\end{tabular}
}
\end{table}

The outcomes of the various approaches are summarized in Table \ref{results}, providing a comprehensive overview. For pure odometry, the average translational error amounted to 0.28m, accompanied by an average rotational error of 0.066rad. As depicted in Fig. \ref{figure:track_raw}, we observe a noticeable drift of the odometry from the ground truth as time progresses.

\begin{figure}[h]
  \centering
  \subfigure[Relative translational error between the static robot and the moving object.]{
\label{figure:odom_level_tran}
        \includegraphics[width=0.45\textwidth]{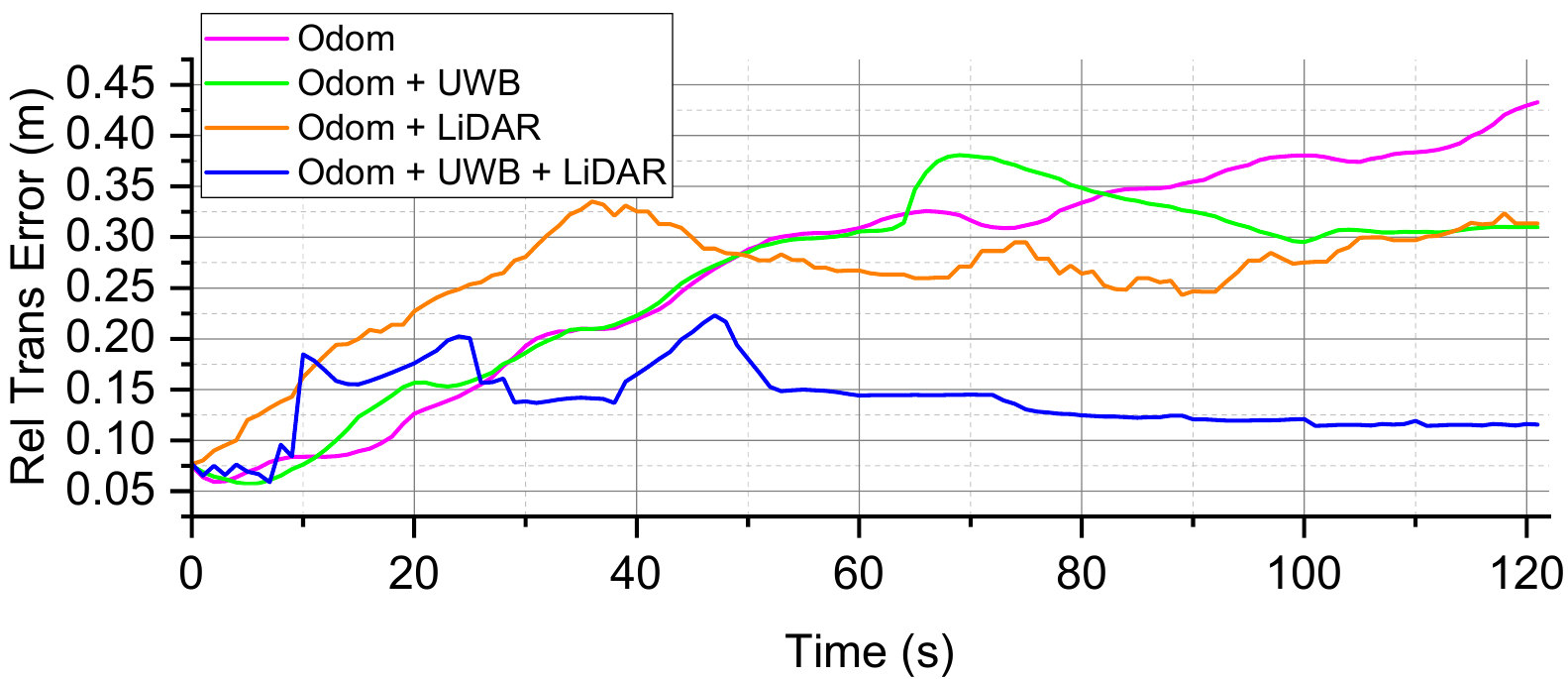}
        }
        \subfigure[Relative rotational error between the static robot and the moving object.]{
\label{figure:odom_level_rot}
        \includegraphics[width=0.45\textwidth]{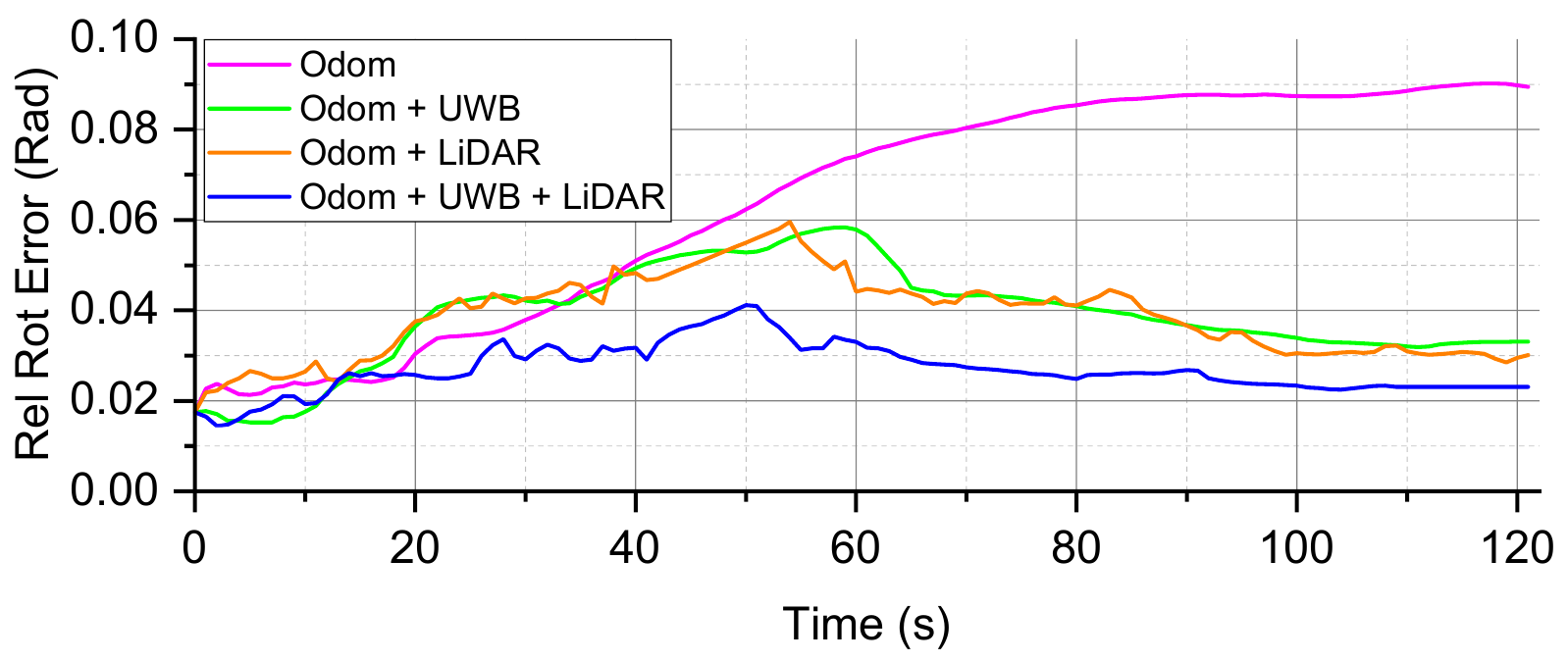}
        }
   \caption[]
{Experimental evaluation of the relative translational and rotational error between the static robot and the moving object with the proposed approach (with moving direction and rejection) over time.}
\label{figure:odom_level_error}
\vspace{-0.3cm}
\end{figure}

The Odom + UWB approach (UWB ranging localization) in Figure \ref{figure:track_pcm} improved the error caused by drift, with a 0.25m average translational error and a 0.038rad rotational error. Although introducing UWB ranging as a constraint improved translational and rotational accuracy, UWB ranging measurements between the static robot and the moving object in non-line-of-sight scenarios will be longer than the actual distance. Fortunately, the data from LiDAR attached to the static robot could help further improve the localization accuracy and compensate for this error.

The Odom + LiDAR approach (fusion of odom and LiDAR without odometry) which introduces LiDAR measurements in the localization algorithm through our object-identification module produced comparable results to UWB ranging localization with a 0.26m average translational error and a 0.039rad rotational error. However, the localization accuracy can be further improved by adding an additional constraint through UWB ranging measurements. Our proposed approach builds on the UWB-ranging localization approach. We evaluated our proposed method without moving direction, without outlier rejection, and with both moving direction and outlier rejection to show the importance of the moving direction estimation and rejecting outliers. With moving direction estimations and the outlier rejection mechanism, the proposed approach produced the best results with a 0.14m average translational error and a 0.026rad rotational error. We compare the estimated moving direction of the moving object from our moving object identification module with the current estimated orientation from PGO. If the difference is within a threshold of 0.3rad, we set a high value for the moving direction in the information matrix. In addition, the average translational and rotational error of the different approaches over time was also evaluated, as shown in Figure \ref{figure:odom_level_error}. We show that our proposed approach maintains a consistent translational and rotational error over time, with a maximum of 0.225m and 0.041rad, respectively.

\subsubsection{Impact of Different Object Moving Direction Information Matrix Values}

\begin{figure*}
  \centering
  \subfigure[Moving robot and moving object ground truth trajectory]{
\label{figure:moving_gt}
        \includegraphics[width=0.283\textwidth]{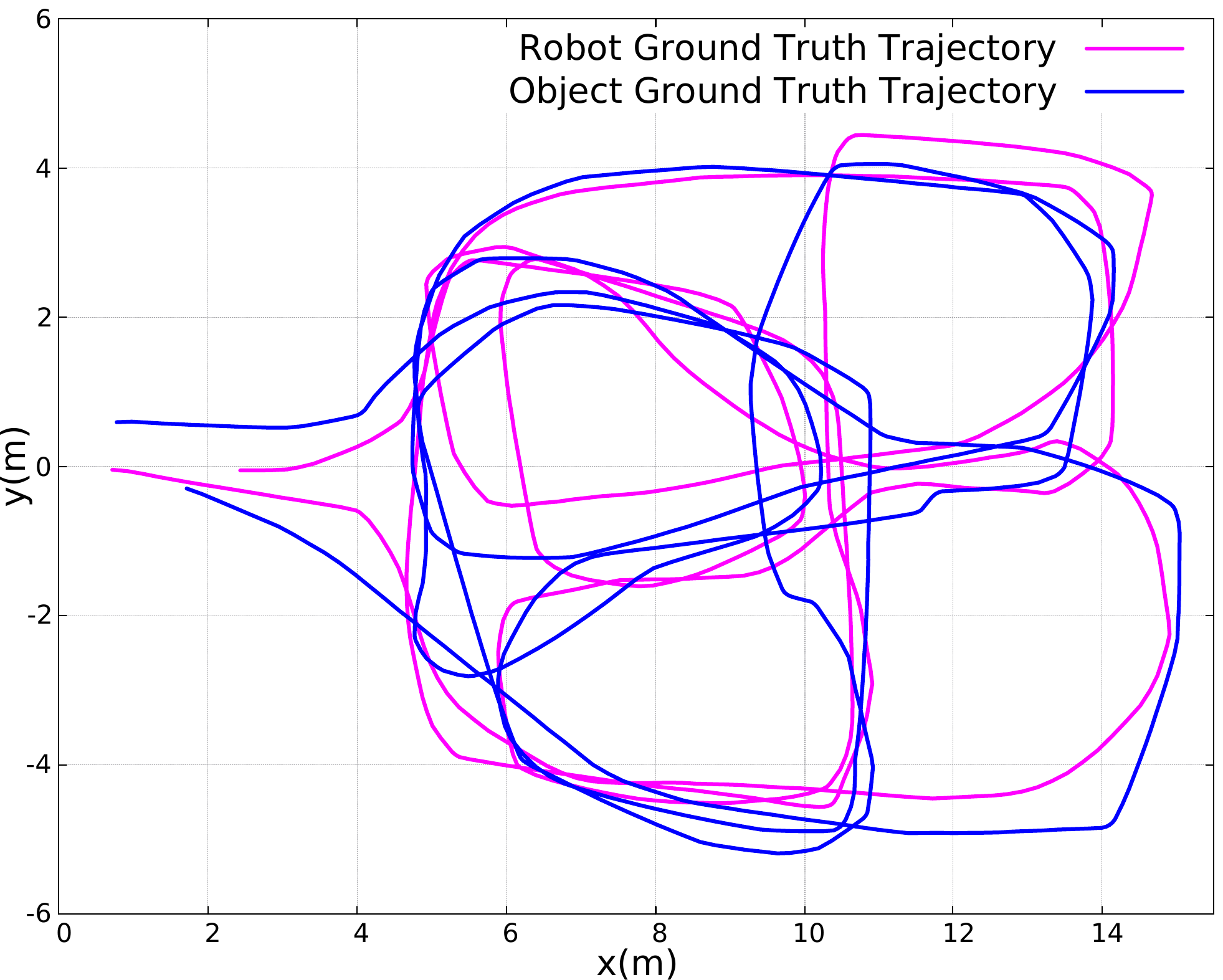}
        }\hspace{0.2in}\subfigure[Moving robot and moving object trajectory with UWB ranging localization]{
\label{figure:moving_uwb}
        \includegraphics[width=0.283\textwidth]{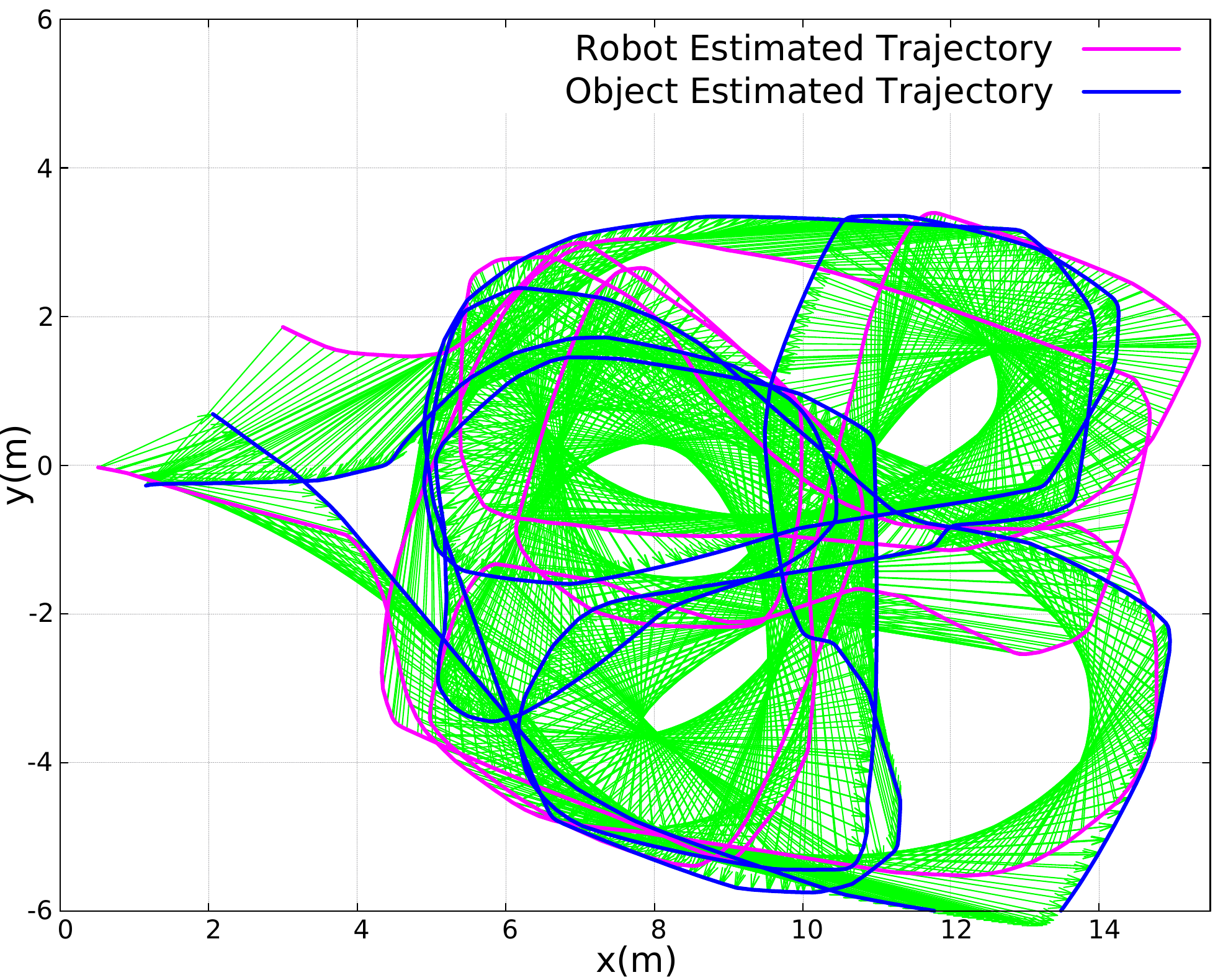}
        }\hspace{0.2in}\subfigure[Moving robot and moving object trajectory with proposed approach (with both moving direction and rejection)]{
\label{figure:moving_uwb_lidar}
        \includegraphics[width=0.283\textwidth]{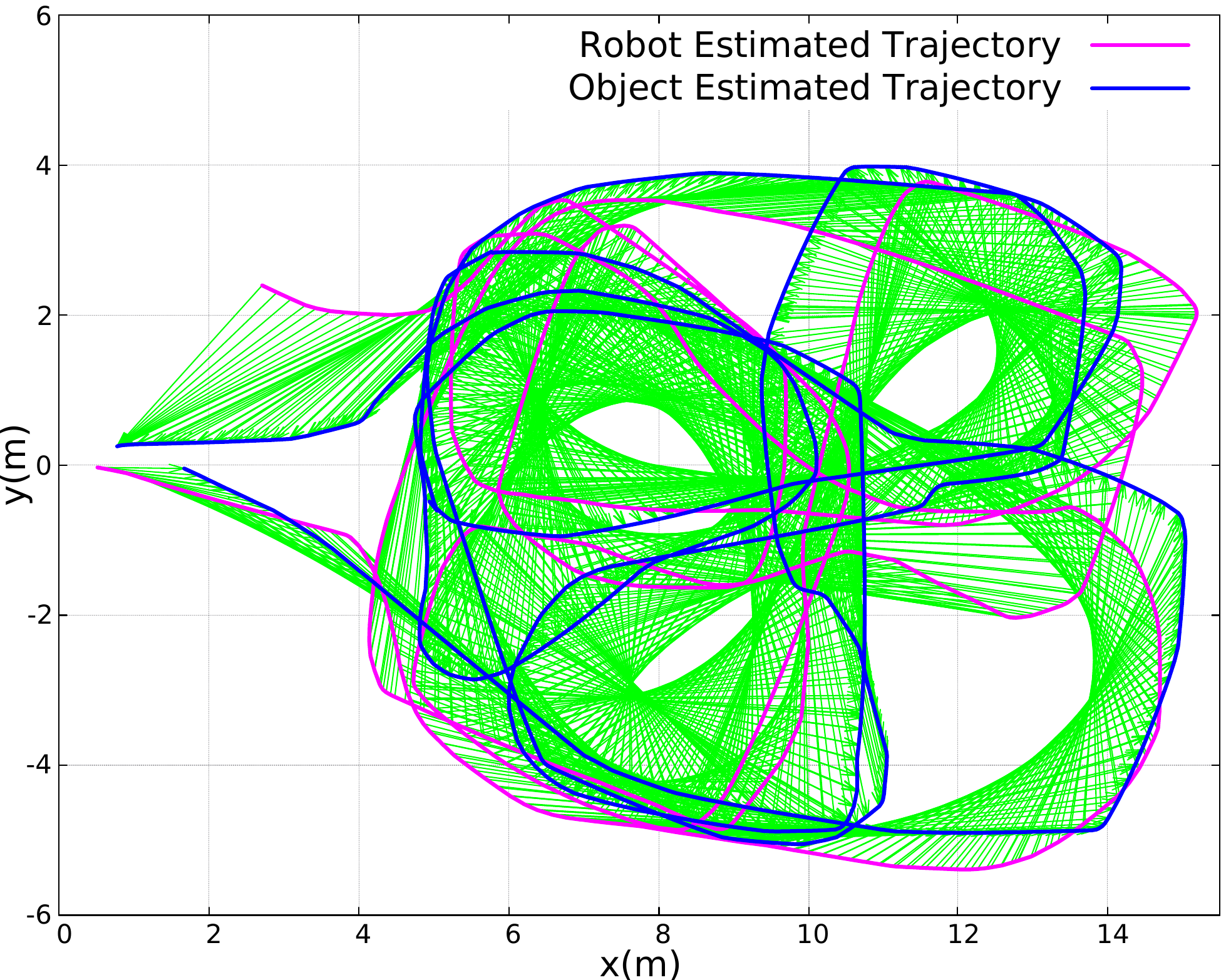}
        }
   \caption[]
{Trajectory of the moving object and moving robot ground truth trajectory, with UWB ranging localization, and with the proposed approach (with moving direction and rejection). The green lines refer to the UWB-ranging constraints between the moving robot and the moving object.}
\label{fig:robot_moving}
\vspace{-0.3cm}
\end{figure*}

\begin{table}[h]
\centering
\caption{Evaluation of different outlier rejection threshold values  $\vartheta$ on the average relative translational error (metres) and relative rotational error (radians) between the static robot and the moving object.}
\label{orientationthreshold}
\resizebox{\linewidth}{!}{%
\begin{tabular}{|c|c|c|} 
\hline
Value $\vartheta$ & Rel. trans error (m) & Rel. rot error (rad) \\ 
\hline
0.1 & 0.28 ± 0.10 & 0.055 ± 0.020 \\ 
\hline
0.2 & 0.19 ± 0.069 & 0.039 ± 0.012 \\ 
\hline
0.3 & 0.14 ± 0.032 & 0.026 ± 0.006 \\ 
\hline
0.4 & 0.17 ± 0.030 & 0.039 ± 0.012 \\ 
\hline
0.5 & 0.23 ± 0.076 & 0.049 ± 0.016 \\
\hline
\end{tabular}
}
\end{table}

\begin{table}[h]
\centering
\caption{Evaluation of different orientation information matrix values $\omega$ on the average relative translational error (metres) and relative rotational error (radians) between the static robot and the moving object.}
\label{orientationinformation}
\resizebox{\linewidth}{!}{%
\begin{tabular}{|c|c|c|} 
\hline
Value $\omega$ & Rel. trans error (m) & Rel. rot error (rad) \\ 
\hline
1 & 0.30 ± 0.15 & 0.063 ± 0.031 \\ 
\hline
10 & 0.24 ± 0.11 & 0.052 ± 0.018 \\ 
\hline
100 & 0.21 ± 0.079 & 0.042 ± 0.016 \\ 
\hline
1000 & 0.19 ± 0.069 & 0.042 ± 0.009 \\ 
\hline
10000 & 0.14 ± 0.032 & 0.026 ± 0.006 \\ 
\hline
100000 & 0.15 ± 0.045 & 0.030 ± 0.009 \\
\hline
\end{tabular}
}
\end{table}
In this experiment, the information matrix for the static robot odometry $\Omega_{R}^{t}$, moving object odometry $\Omega_{O}^{t}$, UWB ranging $\Omega_{R, O}^{t}$, and moving object position $\Omega_{L,\x}^{t}$ constraints are set to 1. We investigated different values $\omega$ for the information matrix of the estimated object moving direction $\Omega_{L, \theta}^{t}$ provided by our object-identification module for PGO. We perform a simple criterion check for the reliability of the estimated moving direction from our proposed approach by comparing it with the estimated object orientation from PGO. We observe that if the moving direction estimated by our proposed approach and the current orientation estimation from PGO is within a threshold of 0.3rad based on experimentation shown in Table \ref{orientationthreshold}, setting the value for the object moving direction information matrix $\Omega_{L, \theta}^{t}$ to 10000 increases the localization accuracy most. We show improved localization accuracy as the value for the object moving direction information matrix $\Omega_{L,\theta}^{t}$ increases up to 10000. However, when the values go above 10000, it causes the localization accuracy to get worse slowly. 
\subsection{Experiments with a Moving Robot and a Moving Object}
\begin{table}[h!]
\centering
\caption{Evaluation of different approaches based on average relative translational error (metres) and relative rotational error (radians) between the moving robot and the moving object.}
\label{results_moving}
\resizebox{\linewidth}{!}{%
\begin{tabular}{|c|c|c|}
\hline
Approach                                                                                        & Rel. trans error (m) & Rel. rot error (rad) \\ \hline
Pure Odom                                                                                       & 0.79 ± 0.34          & 0.28 ± 0.100         \\ \hline
Odom + UWB                                                                                      & 0.52 ± 0.17          & 0.25 ± 0.065         \\ \hline
\begin{tabular}[c]{@{}c@{}}Odom + LiDAR\\ (with moving direction\\ and rejection)\end{tabular}  & 0.52 ± 0.30          & 0.021 ± 0.064        \\ \hline
\begin{tabular}[c]{@{}c@{}}Odom + UWB + LiDAR\\ (with moving direction\\ and rejection)\end{tabular} & 0.45 ± 0.032         & 0.16 ± 0.050         \\ \hline
\end{tabular}
}
\end{table}

\begin{figure}[h]
  \centering
  \subfigure[Relative translational error between the moving robot and the moving object.]{
\label{figure:odom_level_tran_moving}
        \includegraphics[width=0.45\textwidth]{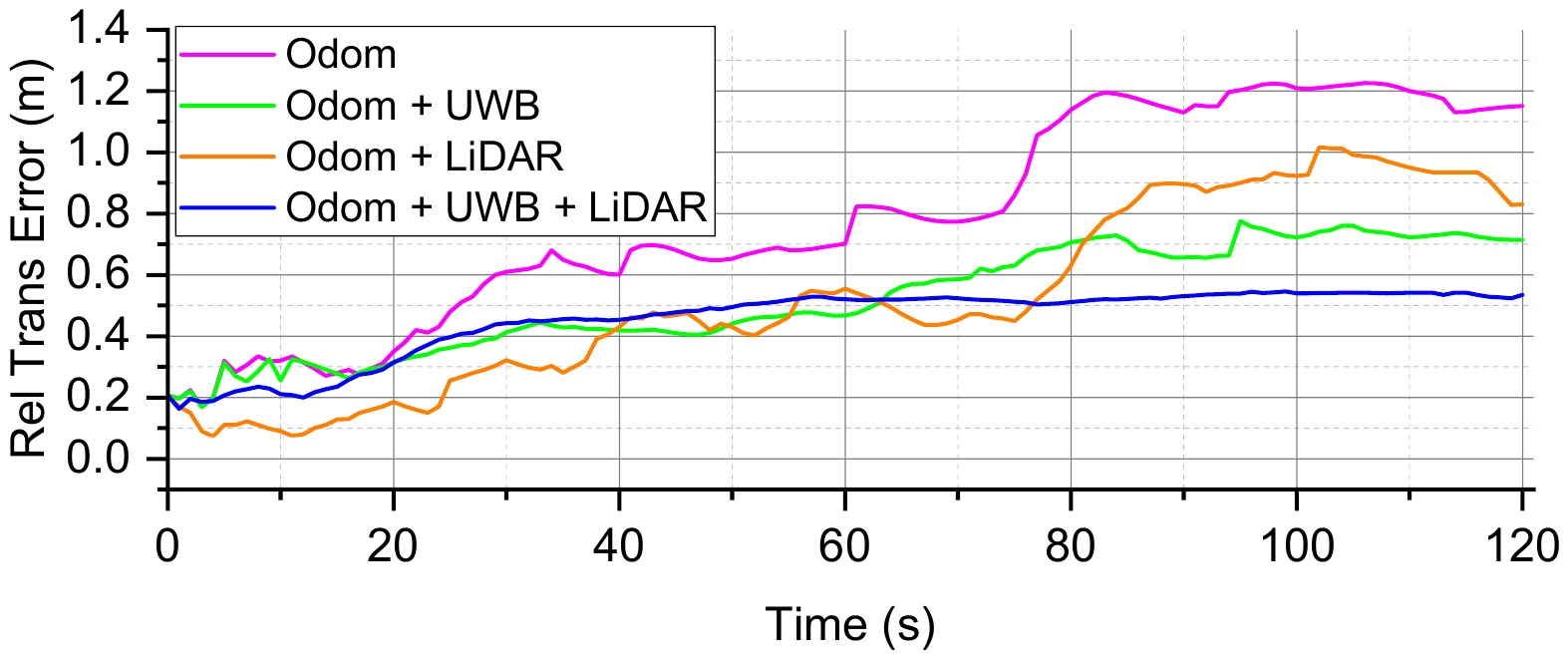}
        }
        \subfigure[Relative rotational error between the moving robot and the moving object.]{
\label{figure:odom_level_rot_moving}
        \includegraphics[width=0.45\textwidth]{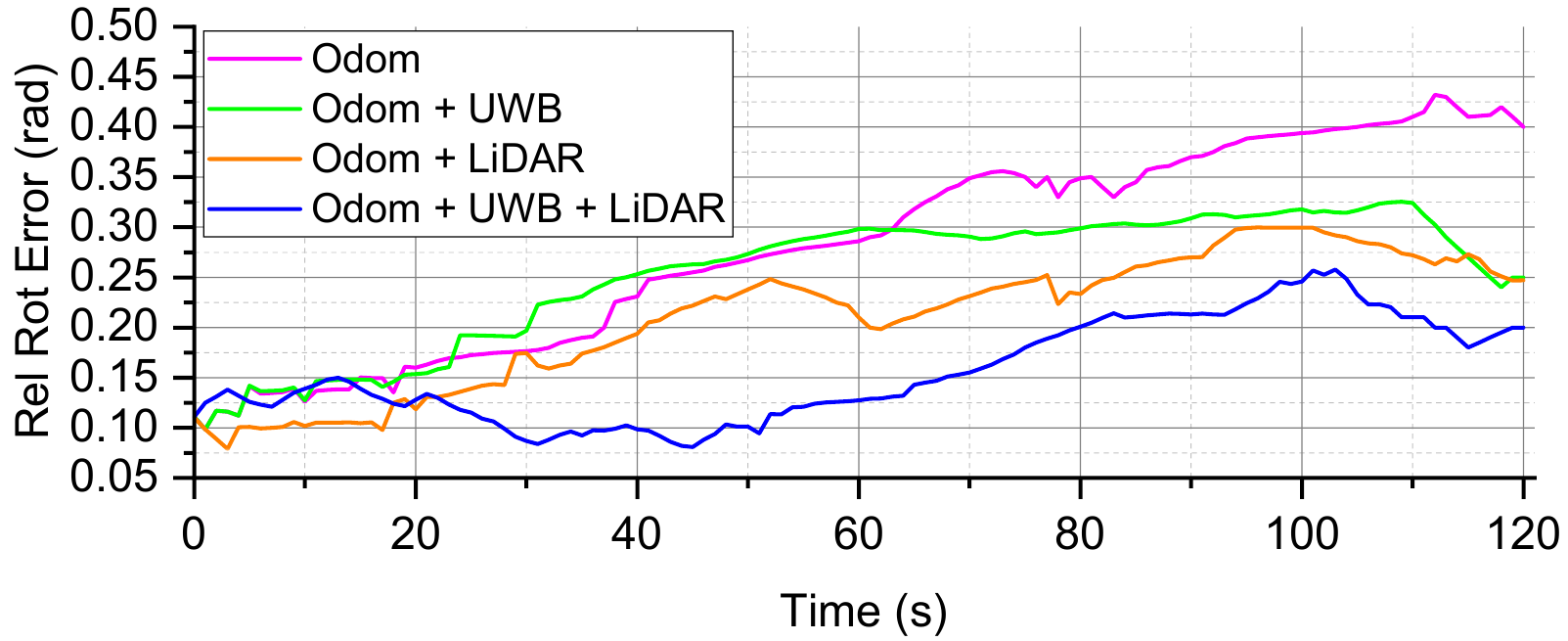}
        }
   \caption[]
{Experimental evaluation of the relative translational and rotational error between the moving robot and the moving object with the proposed approach (with moving direction and rejection) over time.}
\label{figure:odom_level_error_moving}
\vspace{-0.3cm}
\end{figure}
This section presents the experimental results demonstrating the proposed approach with a moving robot and a moving object. The moving robot and the moving object were manually controlled to move along different paths. The best values for $\vartheta$ of 0.3 and $\omega$ of 10000 from Section IV-B2 were used for evaluating the proposed method against pure odometry, UWB ranging localization, and the fusion of odometry and LiDAR measurements only. Figure \ref{fig:robot_moving} visualizes the ground truth trajectory, the estimated trajectory with UWB ranging localization, and also the estimated trajectory with the proposed approach for the moving robot and the moving object. 

We show in Figure \ref{figure:odom_level_error_moving} that our proposed approach produces significant improvements in the relative translational and rotational accuracy between the moving robot and the moving object compared to pure odometry, the conventional UWB ranging localization, and fusion of odometry and LiDAR measurements only. Furthermore, we highlight the improvements of our proposed approach in Table \ref{results_moving}, where the proposed approach improved the conventional UWB ranging localization by 13.5\% and 36\% in the relative translation and rotation error respectively. 

\section{Conclusions}
We proposed an approach for moving object localization using UWB ranging, odometry, and LiDAR measurements between a moving object and a  robot in unknown environments. Our approach consists of three modules that identify the moving object's position using LiDAR, estimate the moving object's moving direction and reject outlier moving direction estimations, and perform PGO to estimate the moving object's position. The proposed approach was verified between a  robot and a moving object in an indoor environment of 16m$\times$12m with obstacles. The results showed that the proposed approach achieved an average localization accuracy of 0.14m in translation and 0.026rad in rotation, which significantly improves accuracy compared to the conventional UWB ranging localization in an environment with one static robot and one moving object. We also showed the importance of moving direction estimations and an outlier rejection mechanism to discard suspicious moving direction estimates from our object-identification module. Additionally, the proposed approach was tested with a moving robot and a moving object which produced significant improvements compared to the conventional UWB ranging localization. In future works, we will extend the work with multiple robots identifying multiple moving objects given that the moving objects can provide odometry. Another research direction is to apply the proposed approach for autonomous moving object-following tasks using a moving robot.

\bibliographystyle{IEEEtran}
\bibliography{bib}

\end{document}